\documentclass[letterpaper, 10 pt, conference]{ieeeconf}  %

\IEEEoverridecommandlockouts                              %

\usepackage{amssymb}
\usepackage{amsmath}
\usepackage{amsfonts}       %
\usepackage{booktabs}       %
\usepackage{balance}
\usepackage[utf8]{inputenc} %
\usepackage[T1]{fontenc}    %
\usepackage[pdftex, pdfstartview={FitV}, pdfpagelayout={TwoColumnLeft},bookmarksopen=true,plainpages = false, colorlinks=true, linkcolor=black, citecolor = black, urlcolor = black,filecolor=black , pagebackref=false,hypertexnames=false, plainpages=false, pdfpagelabels ]{hyperref}
\usepackage{url}            %
\usepackage{nicefrac}       %
\usepackage{microtype}      %
\usepackage[font=scriptsize]{caption}
\usepackage{subcaption}
\usepackage{graphicx}
\usepackage{epstopdf} %
\usepackage{mathtools}
\usepackage{marginnote}
\usepackage[dvipsnames]{xcolor}
\usepackage{float}
\usepackage[capitalize]{cleveref}
\usepackage[]{algorithmic}
\usepackage{makecell}
\usepackage{multirow}
\usepackage{paralist}
\floatstyle{boxed} \floatname{algorithm}{Algorithm}
\newfloat{algorithm}{t}{loa}[section]
\usepackage{xcolor}
\usepackage[nolist,nohyperlinks]{acronym}
\usepackage[sort,compress]{cite}
\usepackage{soul}

\usepackage{amsmath}
\usepackage{color}
\usepackage{soul}
\usepackage{xspace}
\usepackage[capitalize]{cleveref}
\usepackage{dsfont}
\usepackage{bm}
\usepackage{bbm}
\usepackage{nicefrac}  %
\usepackage{amsfonts}
\usepackage{amssymb}
\usepackage{array}
\usepackage{mathtools}
\usepackage{caption}
\usepackage{subcaption}
\usepackage{fontawesome}
\usepackage{thmtools}
\usepackage{thm-restate}
\usepackage{lipsum}
\usepackage{pifont}
\usepackage{makecell}
\usepackage{multirow}
\usepackage{afterpage}
\usepackage{tabularx} 
\usepackage{url}
\usepackage[normalem]{ulem}
\usepackage{wrapfig}  %
\usepackage{textcomp}

\Crefname{figure}{Figure}{Figures}
\crefname{figure}{Figure}{Figures}
\crefname{table}{Table}{Tables}
\crefformat{table}{Table~#1}
\crefformat{equation}{Equation (#1)}
\Crefformat{equation}{Equation (#1)}
\crefformat{algorithm}{Algorithm~#1}
\crefformat{appendix}{Appendix~#1}
\Crefformat{appendix}{Appendix~#1}
\crefformat{appendices}{Appendices~#1}
\Crefformat{appendices}{Appendices~#1}

\newcommand\sdots{\hbox to 1em{.\hss.\hss.}} %
\DeclareMathAlphabet\mathbfcal{OMS}{cmsy}{b}{n}  %

\newif\ifcomments
\commentstrue

\newcommand{\vbf}[2]{\prescript{}{}{\mathbf{#2}}}
\newcommand{\vbs}[2]{\prescript{}{}{\boldsymbol{#2}}}
\newcommand{\vbsd}[2]{\prescript{}{}{\dot{\boldsymbol{#2}}}}
\newcommand{\vbfd}[2]{\prescript{}{}{\dot{\mathbf{#2}}}}

\newcommand{\xdes}{\mathbf{x}^\text{des}}    %
\newcommand{\xsafe}{\bar{\mathbf{x}}}   %
\newcommand{\usafe}{\bar{\mathbf{u}}}   %
\newcommand{\x}{\mathbf{x}}    %

\begin{acronym}
\acro{OC}{Optimal Control}
\acro{LQR}{Linear Quadratic Regulator}
\acro{MAV}{Micro Aerial Vehicle}
\acro{GPS}{Guided Policy Search}
\acro{UAV}{Unmanned Aerial Vehicle}
\acro{MPC}{model predictive control}
\acro{RTMPC}{robust tube model predictive controller}
\acro{DNN}{deep neural network}
\acro{BC}{Behavior Cloning}
\acro{DR}{Domain Randomization}
\acro{SA}{Sampling Augmentation}
\acro{IL}{Imitation learning}
\acro{DAgger}{Dataset-Aggregation}
\acro{MDP}{Markov Decision Process}
\acro{VIO}{Visual-Inertial Odometry}
\acro{VSA}{Visuomotor Sampling Augmentation}
\acro{CoM}{Center of Mass}
\acro{SD}{Standard Deviation}
\acro{MSE}{Mean Squared Error}
\acro{RMSE}{Root Mean Square Error}
\end{acronym}
\crefformat{equation}{(#2#1#3)}

\title{\LARGE \bf
Robust, High-Rate Trajectory Tracking on Insect-Scale Soft-Actuated Aerial Robots  with Deep-Learned Tube MPC 
}

\author{Andrea Tagliabue$^{1,*}$, Yi-Hsuan Hsiao$^{2,*}$, Urban Fasel$^{3}$, J. Nathan Kutz$^{4}$, Steven L. Brunton$^{5}$, \\ YuFeng Chen$^{2}$, Jonathan P. How$^{1}$%
\thanks{*equal contribution}%
\thanks{Work funded in part by the AFOSR MURI FA9550-19-1-0386 and the National Science Foundation (FRR-2202477).}%
\thanks{$^{1}$ Department of Aeronautics and Astronautics, Massachusetts Institute of Technology. \tt\{atagliab, jhow\}@mit.edu}%
\thanks{$^{2}$ Department of Electrical Engineering and Computer Science, Massachusetts Institute of Technology. \tt\{yhhsiao, yufengc\}@mit.edu}%
\thanks{$^{3}$ Department of Aeronautics, Imperial College London. \hspace{\fill} \tt{u.fasel@imperial.ac.uk}}%
\thanks{$^{4}$ Department of Applied Mathematics, University of Washington. \hspace{\fill} \tt{kutz@uw.edu}}%
\thanks{$^{5}$ Department of Mechanical Engineering, University of Washington. \hspace{\fill} \tt{sbrunton@uw.edu}}%
}%

\begin{document}

\maketitle
\thispagestyle{empty}
\pagestyle{empty}

\begin{abstract}
Accurate and agile trajectory tracking in sub-gram \acp{MAV} is challenging, as the small scale of the robot induces large model uncertainties, demanding robust feedback controllers, while the fast dynamics and computational constraints prevent the deployment of computationally expensive strategies.  In this work, we present an approach for agile and computationally efficient trajectory tracking on the MIT SoftFly \cite{chen2019controlled}, a sub-gram \ac{MAV} ($0.7$ grams). Our strategy employs a cascaded control scheme, where an adaptive attitude controller is combined with a neural network policy trained to imitate a trajectory tracking \ac{RTMPC}. The neural network policy is obtained using our recent work \cite{tagliabue2021demonstration}, which enables the policy to preserve the robustness of \ac{RTMPC}, but at a fraction of its computational cost. 
We experimentally evaluate our approach, achieving position \acp{RMSE} lower than $1.8$ cm even in the more challenging maneuvers, obtaining a $60\%$ reduction in maximum position error compared to \cite{chen2021collision}, and demonstrating robustness to large external disturbances. 
\end{abstract}

\section{Introduction}
\label{sec:introduction}

Flying insects exhibit incredibly agile flight abilities, being capable of performing a flip in only $0.4$ ms \cite{liu2019flies}, flying under large wind disturbances \cite{fuller2014flying, ortega2014hawkmoth,ristroph2010discovering}, and withstanding collisions \cite{dickerson2012mosquitoes, mountcastle2014biomechanical}. Insect-scale flapping-wing \acp{MAV} \cite{ma2013controlled,chen2019controlled, yang2019bee,chukewad2021robofly} have the potential to inherit these robustness and agile flight properties, extending their applications  to tight and narrow spaces that become difficult for larger scale \acp{MAV}  \cite{faa_drone_report, flyability, skydio, d2014guest, voliro}. A key capability needed for the deployment of sub-gram \acp{MAV} in real-world missions is the ability to accurately track desired agile trajectories while being robust to real-world uncertainties, such as collisions and wind disturbances.

\begin{figure}[t]
    \centering
    \includegraphics[width=\columnwidth]{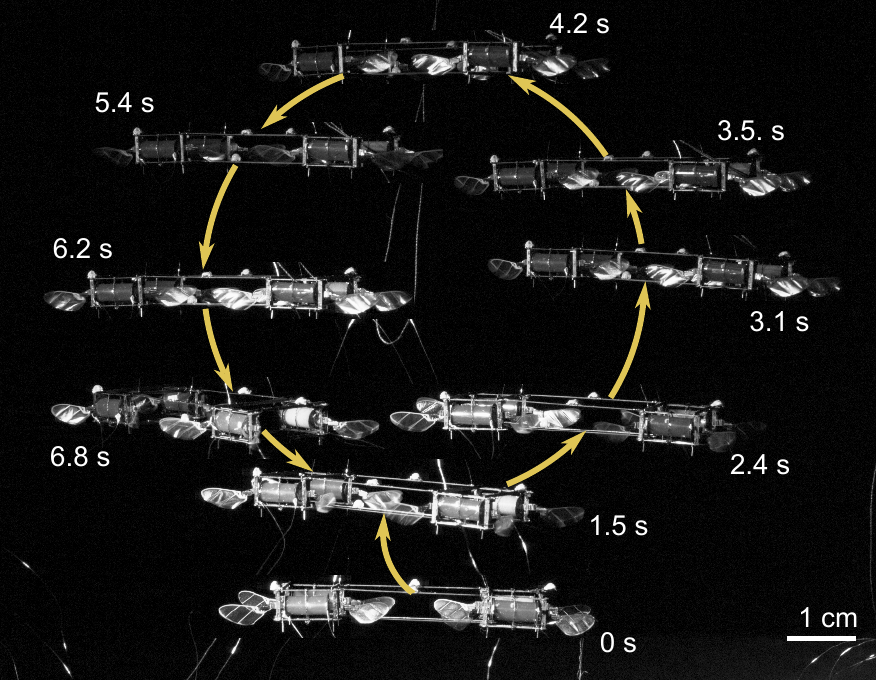}
    \caption{Composite image showing a $7.5$-second flight where the MIT SoftFly \cite{chen2019controlled}, a soft-actuated, insect-scale MAV, follows a vertical circle with $5$ cm radius. The robot is controlled by a neural network policy, trained to reproduce the response of a robust model predictive controller. Thanks to its computational efficiency, the neural network controls the robot at $2$ kHz while running on a small offboard computer.}
    \label{fig:teaser}
    \vskip-5ex
\end{figure}

However, achieving robust, accurate, and agile trajectory tracking in sub-gram \acp{MAV} has significant challenges. First, their exceptionally fast dynamics \cite{chen2021collision} demand high-rate feedback control loops to ensure stability and rapid disturbance rejection, while the small payload limits onboard computation capabilities. Additionally, in order to maximize the lifespan of the robot components, control actions need to be planned in a way that is aware of actuation constraints. For instance, soft dielectric elastomer actuators (DEAs) suffer  dielectric breakdown under a high electric field, posing a hard restraint on the maximum operating voltage \cite{ren2022high}. Furthermore, the lifetime of passively-rotating wing hinges can be substantially extended under moderate control inputs. Lastly, manufacturing imperfection due to the small scale, and hard-to-model unsteady flapping-wing aerodynamics make it difficult to identify accurate models for simulation and control. Existing sub-gram \acp{MAV} have demonstrated promising agile flight capabilities \cite{chen2021collision, School2021Apr, chirarattananon2013adaptive}, but none of the existing controllers \textit{explicitly} account for environment and model uncertainties, and for actuation constraints and usage.

An agile trajectory tracking strategy that has found success on larger-scale \acp{MAV} (e.g., palm-sized quadrotors) consists of decoupling position and attitude control via a cascaded scheme, where a fast feedback loop (\textit{inner}) controls the attitude of the \ac{MAV}, while a potentially slower loop (\textit{outer}) tracks the desired trajectory by generating commands for the attitude controller. An \textit{outer loop} controller that enables agile, robust, actuation-aware trajectory tracking is \ac{MPC} \cite{borrelli2017predictive, lopez2019dynamic, lopez2019adaptive, li2004iterative, kamel2017linear, minniti2019whole, williams2016aggressive}. This strategy generates actions by minimizing an objective function that explicitly trades tracking accuracy for actuation usage, taking into account the state and actuation constraints. This is achieved by solving a constrained optimization problem online, where a model of the robot is employed to plan along a predefined temporal horizon by taking into account the effects of future actions. Robust variants of \ac{MPC}, such as \acf{RTMPC} \cite{mayne2005robust, lopez2019adaptive}, can additionally take into account uncertainties (disturbances, model errors) when generating their plans and control actions. This is done by 
employing an auxiliary (ancillary) controller capable of maintaining the system within some distance (``cross-section'' of a tube) from the nominal plan regardless of the realization of uncertainty. 
While \ac{MPC} and \ac{RTMPC} enable impressive performance on complex, agile robots, their computational cost limits the opportunities for onboard, high-rate deployment on computationally constrained platforms.

In this work, we present and experimentally demonstrate a computationally efficient method for accurate, robust, MPC-based trajectory tracking on sub-gram-\acp{MAV}. Our method uses a cascaded control scheme, where the attitude is controlled via the geometric attitude controller in \cite{lee2010geometric}, which presents a large region of attraction (initial attitude error should be $<180 \deg$). This controller is additionally modified with a parametric adaptation scheme, where a torque observer estimates and compensates for the effects of slowly varying torque disturbances. 
Agile, robust, and real-time implementable trajectory tracking is achieved by employing a computationally efficient deep-neural network policy, trained to imitate the response of a \ac{RTMPC}, given a desired trajectory and the current state of the robot. The neural network policy is obtained using our recent imitation learning method \cite{tagliabue2021demonstration}, which uses a high-fidelity simulator and properties of the controller to generate training data. A key benefit of our method \cite{tagliabue2021demonstration} is the ability to train a computationally efficient policy 
in a computationally efficient way (e.g., a new policy can be obtained in a few minutes), greatly accelerating the tuning phase of the neural network controller.
We experimentally evaluate our robust, agile trajectory tracking approach on the MIT sub-gram-\ac{MAV} SoftFly \cite{chen2021collision}, demonstrating that our method can consistently achieve low position tracking error on a variety of trajectories, which include a circular trajectory (\cref{fig:teaser}) and a ramp, while running at $2$ kHz on a Baseline Target Machine, SpeedGoat offboard computer.
We additionally demonstrate that our strategy is robust to large external disturbances, intentionally applied while the robot tracks a given trajectory.

\noindent
In summary, our work presents the following \textbf{contributions}: 
\begin{itemize}
\item We present the first computationally-efficient strategy for robust, MPC-like control of sub-gram \acp{MAV}. Our approach employs a deep-learned neural network policy that is trained to reproduce a trajectory tracking \ac{RTMPC}, leveraging our previous imitation learning work \cite{tagliabue2021demonstration}.

\item We present a cascaded control strategy, where the attitude controller in \cite{lee2010geometric} is modified with a model adaptation method to compensate for the effects of uncertainties.

\item We perform an experimental evaluation on the MIT SoftFly \cite{chen2021collision}, an agile sub-gram \ac{MAV} ($0.7$ g), showing a $60\%$ reduction in maximum trajectory tracking errors over \cite{chen2021collision}, while being real-time implementable ($2$ kHz) on a small computational platform.
\end{itemize}

\section{Robot Design and Model}
\label{sec:robot_design_and_model}

\begin{figure}
    \centering
    \includegraphics[width=\columnwidth]{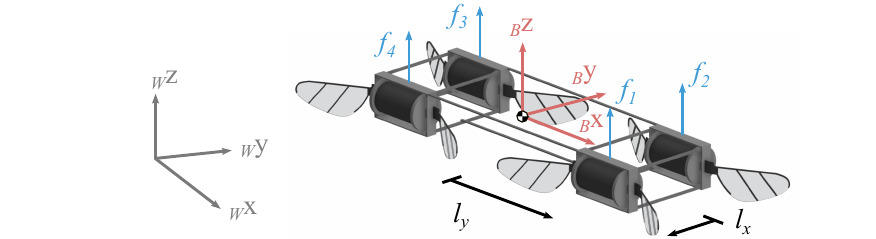}
    \caption{CAD model of sub-gram MAV SoftFly that consists of four soft artificial muscles (DEAs). The inertial reference frame is shown in grey as the body-fixed reference frame is show in red.}
    \label{fig:cad_model}
    \vskip-4ex
\end{figure}

\noindent \textbf{Reference frames.} We consider an inertial reference frame $W=\{\prescript{}{W}{\mathbf{x}}, \prescript{}{W}{\mathbf{y}}, \prescript{}{W}{\mathbf{z}} \}$ and a body-fixed frame $B=\{\prescript{}{B}{\mathbf{x}},\prescript{}{B}{\mathbf{y}}, \prescript{}{B}{\mathbf{z}}\}$ attached to the \ac{CoM} of the robot, as shown in \cref{fig:cad_model}.

\noindent \textbf{Mechanical design.} The sub-gram flapping-wing robot (Figure \ref{fig:cad_model})  consists of four individually-controlled DEAs \cite{ren2022high,chen2019controlled} with newly-designed enhanced-endurance wing hinges \cite{hsiao2023heading}. Unlike natural flying insects that actively control wing stroke and pitch motions \cite{dickinson1999wing}, our robot leverages system resonance (400 Hz) and passive fluid-wing interaction to generate lift forces and support flight. \cite{chen2019controlled,ren2022high}. This design allows each of the four robot modules to generate lift forces without producing significant torques.

\noindent \textbf{Actuation model.}
The voltage inputs to the actuator are controlled to produce desired time-averaged lift forces, and a linear voltage-to-lift-force mapping, $f_i = \alpha_i v_i + \beta_i$, is implemented as previously shown in \cite{chen2019controlled, ren2022high}.
The time-averaged lift force $f_i$ produce by each actuator $i$, with $i = 1, \dots, 4$, is utilized in the model for control purpose since the wing inertia is order of magnitude smaller than that of the robot thorax. These forces can be mapped to the total torque produced by the actuators $\tau_{\text{cmd},x}$ and $\tau_{\text{cmd}, y}$ (around $\prescript{}{B}{\mathbf{x}}$ and $\prescript{}{B}{\mathbf{y}}$, respectively) and $f_\text{cmd}$ as the total thrust force on $\prescript{}{B}{\mathbf{z}}$ via a linear mapping (\textit{mixer} or \textit{allocation} matrix) $\mathcal{A}$: 
\begin{equation}
\label{eq:allocation_matrix}
    \begin{bmatrix}
    f_\text{cmd} \\
    \tau_{\text{cmd},x} \\
    \tau_{\text{cmd},y} \\ 
    \end{bmatrix}
    = \mathcal{A}
    \begin{bmatrix}
    f_1 \\
    \vdots \\
    f_4
    \end{bmatrix}, 
    \hspace{2pt}
    \mathcal{A} =     
    \begin{bmatrix}
    1 & 1 & 1 & 1 \\
    -l_y & l_y & l_y & -l_y \\
    -l_x & -l_x & l_x & l_x \\
    \end{bmatrix},
\end{equation}
where $l_x, l_y$ represent the distance of the actuators from the \ac{CoM} of the robot, as shown in \cref{fig:cad_model}. This configuration 
 (actuators' placement and near-resonant-frequency operating condition) does not generate controlled torques with respect to the body $z$-axis. %

\noindent \textbf{Translational and rotational dynamics.}
The \ac{MAV} is modeled as a rigid body with six degrees of freedom, with mass $m$ and diagonal inertia tensor $\mathbf{J}$, subject to gravitational acceleration $g$. The following set of Newton-Euler equations describes the robot's dynamics: 
\begin{equation}
    \begin{split}
        m\vbfd{W}{v} = & f_\text{cmd} \mathbf{R} \prescript{}{B}{\mathbf{z}} - mg\prescript{}{W}{\mathbf{z}} + \vbf{W}{f}_\text{drag} + \vbf{W}{f}_\text{ext}, \\
        \mathbf{J} \vbsd{B}{\omega} = & -\vbs{B}{\omega} \times \mathbf{J} \vbs{B}{\omega} + \vbs{B}{\tau}_\text{cmd} + \vbs{B}{\tau}_\text{drag} + \vbs{B}{\tau}_\text{ext},\\
        \vbfd{W}{p} = & \vbf{W}{v}, \\
        \dot{\mathbf{R}} = & \mathbf{R}\vbs{B}{\omega}^\wedge. \\
    \end{split}
    \label{eq:mav_dynamic_model}
\end{equation}
Position $\mathbf{p}\in\mathbb{R}^3$ and velocity $\mathbf{v} \in \mathbb{R}^3$ are expressed in $W$; a rotation matrix $\mathbf{R}\in SO(3)$ defines the attitude, and the angular velocity $\vbs{B}{\omega}$ is expressed in $B$. 
We assume that the dynamics are affected by external force and torque $\vbf{W}{f}_\text{ext} \in \mathbb{R}^3$ and $\vbs{B}{\tau}_\text{ext} \in \mathbb{R}^3$, capturing the effects of unknown disturbances, such as the forces/torques applied by the power tethers, imperfections in the assembly and mismatches of model parameters (e.g, mass). Assuming no wind in the environment, we also include an isotropic drag force 
$\vbf{W}{f}_\text{drag} = - c_{Dv} \vbf{W}{v}$ and torque $\vbs{B}{\tau}_\text{drag}=-c_{D\omega}\vbs{B}{\omega}$, with $c_{Dv} > 0, c_{D\omega}>0$.

\section{Flight Control Strategy}
\label{sec:flight_control}
\begin{figure}
    \centering
    \includegraphics[width=\columnwidth, trim={175, 348, 160, 85}, clip]{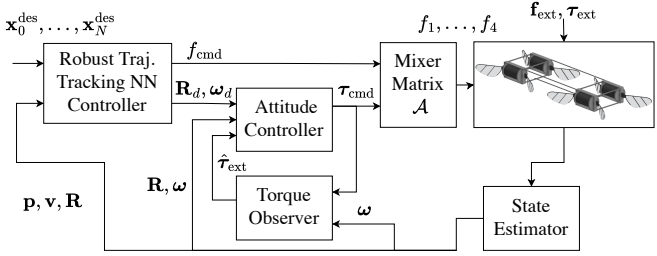}
    \caption{Cascaded control diagram for the proposed robust trajectory tracking control strategy. The robust trajectory tracking neural network (NN) controller constitutes the \textit{outer} loop, and its task is to track a desired trajectory $\mathbf{x}_0, \dots, \mathbf{x}_N$ by generating setpoints $\mathbf{R}_d$, $\boldsymbol{\omega}_d$ for the cascaded attitude controller. The attitude controller and the torque observer constitute the \textit{inner} loop. Thanks to the robustness and adaptive properties of the \textit{outer} and \textit{inner} loops, our approach can withstand the effects of the external disturbances $\boldsymbol{\tau}_\text{ext}, \mathbf{f}_\text{ext}$.}
    \label{fig:control_diagram}
\end{figure}
We decouple trajectory tracking and attitude control via a cascaded scheme, as shown in the control diagram in \cref{fig:control_diagram}. Given an $N+1$-step reference trajectory $\mathbf{x}_0^\text{des}, \dots, \mathbf{x}_N^\text{des}$, a trajectory tracking controller generates desired thrust $f_\text{cmd}$, attitude $\mathbf{R}_d$ and angular velocity $\boldsymbol{\omega}_d$ setpoints. A nested attitude controller then tracks the attitude commands by generating a desired torque $\boldsymbol{\tau}_\text{cmd}$ and by leveraging the estimated torque disturbance $\hat{\boldsymbol{\tau}}_\text{ext}$ provided by a torque observer. The commands $\boldsymbol{\tau}_\text{cmd}, f_\text{cmd}$ are converted (in the Mixer Matrix) to desired mean lift forces $f_1, \dots, f_4$ using the Moore-Penrose inverse $\mathcal{A}^\dagger$ of \cref{eq:allocation_matrix}.

In the following paragraphs, we describe, first, the adaptive attitude controller (\cref{sec:attitude_controller}). Then, we present the computationally expensive trajectory-tracking \ac{RTMPC} (\cref{sec:robust_tube_mpc_overview}) and, last, the computationally efficient procedure to generate the computationally efficient, robust neural network tracking policy (\cref{sec:policy_learning_procedure}) used to control the real robot.

\subsection{Attitude Control and Adaptation Strategy}
\label{sec:attitude_controller}

\textbf{Attitude control law.} The control law employed to regulate the attitude of the robot is based on the Geometric attitude controller in \cite{lee2010geometric}:
\begin{equation}
\begin{split}
    \vbs{B}{\tau}_\text{cmd} = & -\mathbf{K}_R \mathbf{e}_R - \mathbf{K}_\omega \mathbf{e}_\omega + \vbs{B}{\omega} \times \mathbf{J} \vbs{B}{\omega}
    \\ & - \mathbf{J}(\boldsymbol{\omega}^\wedge \mathbf{R}^\top \mathbf{R}_d \boldsymbol{\omega}_d - \mathbf{R}^\top \mathbf{R}_d \vbsd{B}{\omega}_d) - \hat{\boldsymbol{\tau}}_\text{ext},
    \label{eq:attitude_control}
\end{split}
\end{equation}
where $\mathbf{K}_R, \mathbf{K}_\omega$ of size $3 \times 3$ are diagonal matrices, tuning parameters of the controller, and the attitude error $\mathbf{e}_R$ and its time derivative $\mathbf{e}_\omega$ are defined as in \cite{lee2010geometric}: 
\begin{equation}
    \mathbf{e}_R = \frac{1}{2}(\mathbf{R}_d^\top \mathbf{R} - \mathbf{R}^\top \mathbf{R}_d)^{\vee}, \hspace{2pt} \mathbf{e}_\omega = \vbs{B}{\omega} - \mathbf{R}^\top \mathbf{R}_d \vbs{B}{\omega}_d.
\end{equation}
The symbol $(\mathbf{r}^\wedge)^\vee = \mathbf{r}$ denotes the operation transforming a $3 \times 3$ skew-symmetric matrix $\mathbf{r}^\wedge$ in a vector $\mathbf{r} \in \mathbb{R}^3$.
Different from \cite{lee2010geometric}, we assume $\vbsd{B}{\omega}_d = \mathbf{0}_{3}$ to avoid taking derivatives of potentially discontinuous angular velocity commands; we additionally augment \cref{eq:attitude_control} with the adaptive term $\hat{\boldsymbol \tau}_\text{ext}$, which is computed via a torque observer. We note that only the first two components of $\vbs{B}{\tau}_\text{cmd}$ are used for control, as the actuators cannot produce torque $\tau_{\text{cmd},z}$.

\noindent \textbf{Torque observer.} %
We compensate for the effects of uncertainties in the rotational dynamics by estimating torque disturbances $\vbs{B}{\tau}_\text{ext}$ via a steady state Kalman filter. These disturbances are assumed to be slowly varying when expressed in the body frame $B$. The state of the filter is $\mathbf{x}_o = [\vbs{B}{\omega}, \vbs{B}{\tau}_\text{ext}]^\top$.
We assume that the rotational dynamics, employed to compute the prediction (\textit{a priori}) step, evolve according to:
\begin{equation}
    \vbsd{B}{\omega} = \mathbf{J}^{-1} (\mathbf{u}_o + \vbs{B}{\tau}_\text{ext}) + \boldsymbol{\eta}_\omega, \hspace{10pt}
    \prescript{}{}{\dot{\boldsymbol{\tau}}}_\text{ext} = \boldsymbol{\eta}_\tau,
\end{equation}
where $\boldsymbol{\eta}_\omega$, $\boldsymbol{\eta}_\tau$ are assumed to be zero-mean Gaussian noise, whose covariance is a tuning parameter of the filter. Additionally, the prediction step is performed assuming the control input $\mathbf{u}_o = \vbs{B}{\tau}_\text{cmd} - \vbs{B}{\omega} \times \mathbf{J} \vbs{B}{\omega}$, which enables us to take into account the nonlinear gyroscopic effects $\vbs{B}{\omega} \times \mathbf{J} \vbs{B}{\omega}$ while using a computationally efficient linear observer. The measurement update (a \textit{posteriori}) uses angular velocity measurements $\mathbf{z}_o = \vbs{B}{\omega}_m + \boldsymbol{\eta}_m$, assumed corrupted by an additive zero mean Gaussian noise $\boldsymbol{\eta}_m$, whose covariance can be identified from data or adjusted as a tuning parameter.

\subsection{Robust Tube MPC for Trajectory Tracking}
\label{sec:robust_tube_mpc_overview}
In this part,
first we present the hover-linearized vehicle model employed for control design (\cref{sec:model_linearization}). We then present the optimization problem solved by a linear \ac{RTMPC} (\cref{sec:robust_tube_mpc}) and the compensation schemes to account for the effects of linearization (\cref{sec:compensation_scheme}).

\subsubsection{Linearized Model}
\label{sec:model_linearization}
We linearize the dynamics \cref{eq:mav_dynamic_model} around hover using a procedure that largely follows \cite{kamel2017linear, kamel2017model}. 
The key differences are highlighted in the following.

First, for interpretability, we represent the attitude of the \ac{MAV} via the Euler angles yaw $\psi$, pitch $\theta$, roll $\phi$ (\textit{intrinsic} rotations around the $z$-$y$-$x$). The corresponding rotation matrix can be obtained as $\mathbf{R} = \mathbf{R}_{z}(\psi)\mathbf{R}_{y}(\theta) \mathbf{R}_{x}(\phi)$, where $\mathbf{R}_{j}(\alpha)$ denotes a rotation of $\alpha$ around the ${j}$-th axis. Additionally, we express the dynamics \cref{eq:mav_dynamic_model} in a yaw-fixed frame $I$, so that $\prescript{}{I}{\mathbf{x}}$ is aligned with $\prescript{}{W}{\mathbf{x}}$. 
The roll $\prescript{}{I}{\phi}$ and pitch $\prescript{}{I}{\theta}$ angles (and their first derivative $\prescript{}{I}{\varphi}$, $\prescript{}{I}{\vartheta}$) expressed in $I$ can be expressed in $B$ via the rotation matrix $\mathbf{R}_{BI}$: 
\begin{equation}
    \begin{bmatrix}
    \phi \\
    \theta \\
    \end{bmatrix} =
    \mathbf{R}_{BI}
    \begin{bmatrix}
    \prescript{}{I}{\phi} \\
    \prescript{}{I}{\theta} \\
    \end{bmatrix}, \hspace{1pt}
    \mathbf{R}_{BI} =
    \begin{bmatrix}
    \cos(\psi) & \sin(\psi) \\
    -\sin(\psi) & \cos(\psi) \\
    \end{bmatrix}.
\end{equation}
The state $\mathbf{x}$ of the linearized model is chosen to be: 
\begin{equation}
\label{eq:rtmpc_current_state}
\mathbf{x} = [
\vbf{W}{p}^\top, \vbf{W}{v}^\top, \prescript{}{I}\phi, \prescript{}{I}{\theta},  \prescript{}{I}{\delta\phi}_\text{cmd}, \prescript{}{I}{\delta\theta}_\text{cmd}
]^\top, 
\end{equation}
where $\prescript{}{I}{\delta\phi}_\text{cmd}, \prescript{}{I}{\delta\theta}_\text{cmd}$ denote the linearized commanded attitude.
We chose the control input $\mathbf{u}$ to be:
\begin{equation}
\label{eq:rtmpc_control_input}
\mathbf{u} = [ \prescript{}{I}{\varphi}_\text{cmd}, \prescript{}{I}{\vartheta}_\text{cmd},  \delta f_\text{cmd}]^\top, 
\end{equation}
where $\delta f_\text{cmd}$ denotes the linearized commanded thrust,  and $\prescript{}{I}{\varphi}_\text{cmd}$ and $\prescript{}{I}{\vartheta}_\text{cmd}$ are the commanded roll and pitch rates. The choice of $\mathbf{x}$ and $\mathbf{u}$ differs from \cite{kamel2017linear, tagliabue2021demonstration} as the control input consist in the roll, pitch rates rather than $\theta_\text{cmd}, \phi_\text{cmd}$; this is done to avoid discontinuities in $\theta_\text{cmd}, \phi_\text{cmd}$, and to feed-forward angular velocity commands to the attitude controller.

The linear translational dynamics are obtained by linearization of the translational dynamics in \cref{eq:mav_dynamic_model} around hover. 
Linearizing the closed-loop rotational dynamics is more challenging, as they should include the linearization of the attitude controller. Following \cite{kamel2017model}, we model the closed-loop attitude dynamics around hover expressed in $I$ as: 
\begin{equation}
\begin{split}
    \prescript{}{I}{\dot{\theta}} = \frac{1}{\tau_\theta}(k_\theta \prescript{}{I}{\theta}_\text{cmd} - \prescript{}{I}{\theta}), 
    \hspace{5pt}  
    \prescript{}{I}{\dot{\theta}}_\text{cmd} = \prescript{}{I}{\vartheta}_\text{cmd}, \\
    \prescript{}{I}{\dot{\phi}} = \frac{1}{\tau_\phi}(k_\phi \prescript{}{I}{\phi}_\text{cmd} - \prescript{}{I}{\phi}),
    \hspace{5pt}  
    \prescript{}{I}{\dot{\phi}}_\text{cmd} = \prescript{}{I}{\varphi}_\text{cmd},
\end{split}
\end{equation}
where $k_\phi$, $k_\theta$ are gains of the commanded roll and pitch angles, while $\tau_\phi$, $\tau_\theta$ are the respective time constants. These parameters can be obtained via system identification.

Last, we model the unknown external force disturbance $\vbf{W}{f}_\text{ext}$ in \cref{eq:mav_dynamic_model} as a source of  bounded uncertainty, assumed to be $\|\mathbf{f}_\text{ext}\|_\infty < \bar{f}_\text{ext}$. This introduces an additive bounded uncertainty $\mathbf{w} \in \mathbb{W}$, with:
\begin{equation}
\mathbb{W} := \{ \mathbf w = [\boldsymbol{0}_3^\top, \vbf{W}{f}_\text{ext}^\top, \mathbf{0}_4^\top]^\top \hspace{2pt}|\hspace{2pt}   \|\mathbf{f}_\text{ext}\|_\infty < \bar{f}_\text{ext} \}.
\end{equation}

Via discretization with sampling period $T_c$, we obtain the following linear, uncertain state space model:
\begin{equation}
\label{eq:linearized_dynamics}
\mathbf{x}_{t+1} = \mathbf{A} \mathbf{x}_t + \mathbf{B} \mathbf{u}_t + \mathbf{w}_k,
\end{equation}
subject to actuation and state constrains $\mathbb{U} = \{ \mathbf{u} \in \mathbb{R}^3 | \mathbf{u}_\text{min} \leq \mathbf{u} \leq \mathbf{u}_\text{max} \}$, and $\mathbb{X} = \{ \mathbf{x} \in \mathbb{R}^{10} | \mathbf{x}_\text{min} \leq \mathbf{x} \leq \mathbf{x}_\text{max}\}$.

\subsubsection{Controller Formulation}
\label{sec:robust_tube_mpc}
The trajectory tracking \ac{RTMPC} is based on \cite{mayne2005robust}, with the objective function modified to track a desired trajectory. 

\noindent
\textbf{Optimization problem.} At every timestep $t$, \ac{RTMPC} takes as input the current state of the robot $\mathbf x_t$ and a $N+1$-step desired trajectory $\mathbf{X}^\text{des}_t = \{\xdes_{0|t},\dots,\xdes_{N|t}\}$, where $\xdes_{i|t}$ indicates the desired state at the future time $t+i$, as given at the current time $t$. It then computes a sequence of reference states $\mathbf{\bar{X}}_t = \{\xsafe_{0|t},\dots,\xsafe_{N|t}\}$ and actions $\mathbf{\bar{U}}_t = \{\usafe_{0|t},\dots,\usafe_{N-1|t}\}$, that will not cause constrain violations (``safe'') regardless of the realization of $\mathbf{w} \in \mathbb{W}$. This is achieved by solving:
\begin{eqnarray} 
\label{eq:rtmpc_optimization_problem}
    \mathbf{\bar{U}}_t^*, \mathbf{\bar{X}}_t^*
    &=& \underset{\mathbf{\bar{U}}_t, \mathbf{\bar{X}}_t}{\text{argmin}}
        \| \mathbf e_{N|t} \|^2_{\mathbf{P}_x} + 
        \sum_{i=0}^{N-1} 
            \| \mathbf e_{i|t} \|^2_{\mathbf{Q}_x} + 
            \| \mathbf u_{i|t} \|^2_{\mathbf{R}_u} \notag \\
    && \hspace*{-.6in} \text{subject to} \:\:  \xsafe_{i+1|t} = \mathbf A \xsafe_{i|t} + \mathbf B \usafe_{i|t},  \\
     &&\xsafe_{i|t} \in \mathbb{X} \ominus \mathbb{Z}, \:\: \usafe_{i|t} \in \mathbb{U} \ominus \mathbf{K} \mathbb{Z}, \notag\\
     &&\x_t \in \mathbb{Z} \oplus  \xsafe_{0|t},  \notag
\end{eqnarray}
where $\mathbf e_{i|t} = \xsafe_{i|t} - \xdes_{i|t}$ is the tracking error. The positive definite matrices $\mathbf{Q}_x$,
$\mathbf{R}_u$ 
define the trade-off between deviations from the desired trajectory and actuation usage, while $\| \mathbf e_{N|t} \|^2_{\mathbf{P}_x}$ is the terminal cost. $\mathbf{P}_x$ 
and $\mathbf K$ are obtained by solving an infinite horizon optimal control LQR problem using $\mathbf A$, $\mathbf B$, $\mathbf{Q}_x$ and $\mathbf{R}_u$. $\oplus$, $\ominus$ denote set addition (\textit{Minkowski sum}) and subtraction (\textit{Pontryagin difference}).
As often done in practice \cite{kamel2017model}, we do not include a terminal set constraints, but we make sure to select a sufficiently long prediction horizon to achieve recursive feasibility. 

\noindent
\textbf{Tube and ancillary controller.}  A control input for the real system is generated by \ac{RTMPC} via an \textit{ancillary controller}:
\begin{equation}
\label{eq:ancillary_controller}
    \mathbf u_t = \mathbf \usafe^*_{t} + \mathbf K (\x_t - \xsafe^*_{t}),
\end{equation}
where $\usafe^*_{0|t} = \usafe^*_t$ and $\xsafe^*_{0|t} = \xsafe^*_t$. 
This controller ensures that the system remains inside a \textit{tube} (with ``cross-section'' $\mathbb{Z}$) centered around $\xsafe_t^*$ regardless of the realization of the disturbances in $\mathbb{W}$, provided that the state of the system starts in such tube (constraint $\x_t \in \mathbb{Z} \oplus \xsafe_{0|t}$).
The set $\mathbb{Z}$ is a disturbance invariant set for the closed-loop system $\mathbf{A}_K := \mathbf{A + B K}$, satisfying the property that $\forall \mathbf{x}_t \in \mathbb{Z}$, $\forall \mathbf{w}_t \in \mathbb{W}$, $\forall t \in \mathbb{N}^+$, $\mathbf{x}_{t+1} = \mathbf{A}_K \mathbf{x}_t + \mathbf{w}_t \in \mathbb{Z}$. 
In our work, we compute an approximation of $\mathbb{Z}$ by computing the largest deviations from the origin of $\mathbf{A}_K$ by performing Monte-Carlo simulations, with disturbances uniformly sampled from $\mathbb{W}$.  

\subsubsection{Compensation schemes and attitude setpoints}
\label{sec:compensation_scheme}
Following \cite{kamel2017model}, we apply a compensation scheme to the commands: 
\begin{equation}
f_\text{cmd} = \frac{\delta f_\text{cmd} + g}{\cos(\phi)\cos(\theta)}, \hspace{3pt} 
\begin{bmatrix}
\phi_\text{cmd} \\
\theta_\text{cmd} 
\end{bmatrix}
= \frac{g}{f_\text{cmd}}
\begin{bmatrix}
\delta \phi_\text{cmd} \\
\delta \theta_\text{cmd}
\end{bmatrix}.
\end{equation}
This desired orientation $\theta_\text{cmd}$, $\phi_\text{cmd}$ is converted to a desired rotation matrix $\mathbf{R}_d$, setpoint for the attitude controller, setting the desired  yaw angle to the current yaw angle $\psi_\text{cmd} = \psi$. 
Last, the desired angular velocity $\boldsymbol{\omega}_d$ is computed from the desired yaw, pitch, roll rates $\mathbf{q} = [\dot{\psi}_\text{cmd}, \vartheta_\text{cmd}, \varphi_\text{\text{cmd}}]^\top$ via \cite{diebel2006representing} $\boldsymbol{\omega}_d  = \mathbf{E}_{\psi,\theta,\phi} \mathbf{q}$, with
\begin{equation}
    \mathbf{E}_{\psi,\theta,\phi} =
    \begin{bmatrix}
    0 & -\sin(\psi) & \cos(\psi) \sin(\theta) \\
    0 & \cos(\psi) & \sin(\psi) \cos(\theta) \\
    1 & 0 & -\sin(\theta) \\
    \end{bmatrix},
\end{equation}
and assuming the desired yaw rate $\dot{\psi}_\text{cmd} = 0$.

\subsection{Robust Tracking Neural Network Policy}
\label{sec:policy_learning_procedure}
\begin{figure}
    \centering
    \includegraphics[width=\columnwidth, trim={120, 365, 295, 90}, clip]{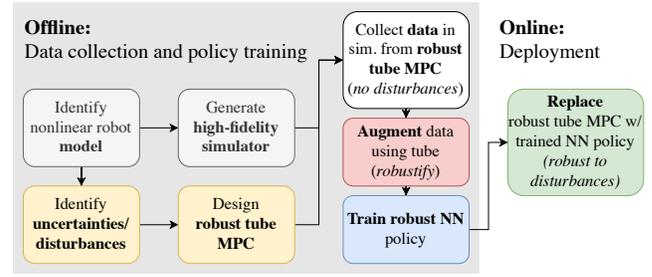}
    \caption{Imitiation learning strategy employed to learn a robust neural network policy from robust tube MPC. Key idea is to collect \ac{RTMPC} demonstrations in simulation and to leverage properties of the tube and of the ancillary controller to augment the collected demonstrations with data that improve the robustness of the trained policy.}
    \label{fig:il_strategy}
    \vskip-2ex
\end{figure}

The overall procedure to generate data needed to train a computationally efficient neural network policy, capable of reproducing the response of the trajectory tracking \ac{RTMPC} described in \cref{sec:robust_tube_mpc_overview}, is summarized in \cref{fig:il_strategy}. 

\noindent \textbf{Policy input-output.} The deep neural network policy that we intend to train is denoted as $\pi_\theta$, with parameters $\theta$. Its input-outputs are the same as the ones of \ac{RTMPC} in \cref{sec:robust_tube_mpc_overview}:
\begin{equation}
\label{eq:nn_policy}
    \mathbf u_t = \pi_\theta(\mathbf{x}_t, \mathbf{X}^\text{des}_t).
\end{equation}

\begin{figure*}[ht!]
    \centering
    \includegraphics[width=\textwidth]{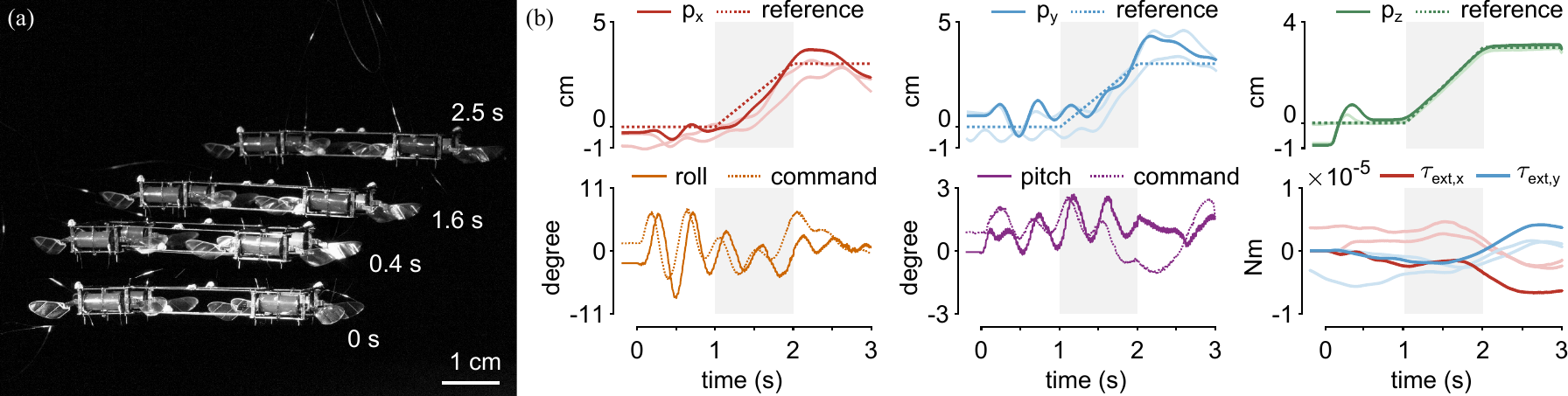}
    \caption{Performance of the proposed flight control strategy in Task 1 (T1), where the robot follows a $3.5$ s ramp trajectory  on $x$-$y$-$z$. The experiment is repeated three times, and the results are included in the shaded lines, with the exception of the roll and pitch angles, for clarity. These results highlight the accuracy of the proposed trajectory tracking error, which achieves a $0.6$ cm position tracking RMSE on $x$-$y$, and $0.2$ cm on $z$.}
    \label{fig:pos_ramp}
    \vskip-2ex
\end{figure*}

\begin{table*}
    \caption{Position Root Mean Squared Errors (RMSE) and Maximum Absolute Errors (MAE) when tracking a ramp (T1), a ramp with disturbances (T2) and a circle (T3). All the values are computed after takeoff ($t > 0.5$ s).}
    \newcolumntype{P}[1]{>{\centering\arraybackslash}p{#1}}
    \centering
    \scriptsize
    \renewcommand{\tabcolsep}{7.5pt}
    \begin{tabular} {||c|c|c|c|c|c|c||c|c||c|c|c|c|c|c||}
    \hline
    \multicolumn{1}{||c|}{} & \multicolumn{6}{c||}{\textbf{T1: Ramp (3 runs, 2.5 s)}} & \multicolumn{2}{c||}{\textbf{T2 (1 run, 7.0 s)}} & \multicolumn{6}{c||}{\textbf{T3: Circle (3 runs, 7.5 s)}} \\
    \multicolumn{1}{||c|}{} & \multicolumn{3}{c|}{RMSE (cm, $\downarrow$)} & \multicolumn{3}{c||}{MAE (cm, $\downarrow$)}& \multicolumn{1}{c|}{RMSE} & \multicolumn{1}{c||}{MAE} & \multicolumn{3}{c|}{RMSE (cm, $\downarrow$)} & \multicolumn{3}{c||}{MAE (cm, $\downarrow$)}  \\
    \multicolumn{1}{||c|}{Axis} & \scriptsize{AVG} & \scriptsize{MIN} & \scriptsize{MAX} & \scriptsize{AVG} & \scriptsize{MIN} & \scriptsize{MAX} & \scriptsize{(cm, $\downarrow$)} & \scriptsize{(cm, $\downarrow$)} & \scriptsize{AVG} & \scriptsize{MIN} & \scriptsize{MAX} & \scriptsize{AVG} & \scriptsize{MIN} & \scriptsize{MAX}\\
    \hline
    \hline
    $x$ & 0.6 & 0.4 & 0.9 & 1.1 & 0.7 & 1.5 & 0.7 & 2.5 & 1.0 & 0.8 & 1.4 & 2.0 & 1.7 & 2.6 \\
    $y$ & 0.8 & 0.7 & 0.9 & 1.5 & 1.3 & 1.6 & 1.0 & 2.1 & 1.5 & 1.3 & 1.8 & 2.8 & 2.5 & 3.1 \\
    $z$ & 0.1 & 0.1 & 0.1 & 0.2 & 0.2 & 0.2 & 0.2 & 0.3 & 0.3 & 0.3 & 0.3 & 0.6 & 0.5 & 0.8 \\
    \hline
    \end{tabular}
    \label{tab:rmse_all}
    \vskip-3ex
\end{table*}

\noindent \textbf{Policy training.}
Our approach, based on our previous work \cite{tagliabue2021demonstration}, consists in the following steps:
\begin{inparaenum}[1)]

\item We design a high-fidelity simulator, where we implement a discretized model (discretization period $T_s$) of the nonlinear dynamics \cref{eq:mav_dynamic_model} and the control architecture consisting of the attitude controller (\cref{sec:attitude_controller}) and \ac{RTMPC} (\cref{sec:robust_tube_mpc_overview}). In simulation, we assume that the \ac{MAV} is not subject to disturbances, setting $\mathbf{f}_\text{ext} = \mathbf{0}_{3}$ and $\boldsymbol{\tau}_\text{ext} = \mathbf{0}_{3}$, and therefore we do not simulate the torque observer. 

\item Given a desired trajectory, we collect a $T+1$-step \textit{demonstration} $\mathcal{T}$ by simulating the entire system controlled by \ac{RTMPC}. At every timestep $t$, we store the input-outputs of \ac{RTMPC}, with the addition of the safe plans $\usafe_t^*, \xsafe_t^*$, obtaining:
\begin{multline}
    \mathcal{T} = ((\mathbf x_0, \mathbf u_0, (\usafe_0^*, \xsafe_0^*), \mathbf{X}^\text{des}_0), \dots \\
    \dots, (\mathbf x_T, \mathbf u_T, (\usafe_T^*, \xsafe_T^*), \mathbf{X}^\text{des}_T)).
\end{multline}

\item We generate the training dataset for the policy. This dataset consists of the input-outputs of the controller obtained from the collected demonstrations, augmented with extra (state, action) pairs $(\mathbf{x}^+, \mathbf{u}^+)$ obtained by uniformly sampling extra states from inside the tube $\mathbf x^+ \in \xsafe_t^* \oplus \mathbb{Z}$, and by computing the corresponding robust control action $\mathbf u^+$ using the ancillary controller:
\begin{equation}
\mathbf u^+ = \mathbf \usafe_t^* +  \mathbf K (\mathbf{x}^+ - \xsafe_t^*).
\label{eq:tubempc_feedback_policy}
\end{equation}
As a result of this data augmentation procedure \cite{tagliabue2021demonstration}, we can generate training data that can compensate for the effects of uncertainties in $\mathbb{W}$. This procedure has also the potential to reduce the time needed for the data collection phase over other existing imitation learning methods, as \cref{eq:tubempc_feedback_policy} can be computed efficiently.

\item The optimal parameters $\theta^*$ for the policy \cref{eq:nn_policy} are then found by training the policy on the collected and augmented dataset, minimizing the \ac{MSE} loss.

\end{inparaenum}

\section{Experimental Evaluation}
\label{sec:evaluation}

The robustness and performance of the described flight controllers are experimentally evaluated on the soft-actuated MIT SoftFly \cite{chen2021collision}. We consider two trajectories of increasing difficulty, a position ramp, where we additionally perturb the \ac{MAV} with external disturbances, and a circular trajectory. 

\noindent \textbf{Experimental setup.}
The flight experiments are performed in a motion-tracking environment equipped with $6$ motion-capturing cameras (Vantage V5, Vicon). The Vicon system provides positions and orientations of the robot, and velocities are obtained via numerical differentiation. The controller runs at $2$ kHz on the Baseline Target Machine (Speedgoat) using the Simulink Real-Time operating system; its commands are converted to sinusoidal signals for flapping motion at $10$ kHz. Voltage amplifiers ($677B$, Trek) are connected to the controller and produce amplified control voltages to the robot. 

\noindent \textbf{Controller and training parameters.} The parameter employed for the controllers are obtained via system identification, also leveraging \cite{brunton2016discovering, fasel2022ensemble}, or via an educated guess. The \ac{RTMPC} has a $1$-second long prediction horizon, with $N=50$ and $T_c=0.02$; we set $\bar{f}_\text{ext}$ to correspond to $15\%$ of the weight force acting on the robot.
State and actuation constraints capture safety and actuation limits of the robot (e.g., max actual/commanded roll/pitch $<25 \deg$,  $\|\delta f_\text{cmd}\| < 80\% mg$).
The employed policy is a $2$-hidden layers, fully connected neural network, with $32$ neurons per layer. Its input size is $310$ (current state, and reference trajectory containing desired position, velocity across the prediction horizon $N$), and its output size is $3$. During the experiments, we slightly tune the parameters of the controllers (e.g., $\mathbf{Q}_x, \mathbf{R}_u, \mathbf{K}_R, \mathbf{K}_\omega$). We train a policy for each type of trajectory (ramp, circle), using the ADAM optimizer, with a learning rate $\eta=0.001$, for $15$ epochs. Data augmentation is performed by extracting from the tube $200$ extra state-action samples per timestep. Training each policy takes about $1$ minute (for $T=350$ steps) on a Intel i9-10920 ($12$ cores) with two Nvidia RTX 3090 GPUs.

\begin{figure*}[ht!]
    \centering
    \includegraphics[width=\textwidth]{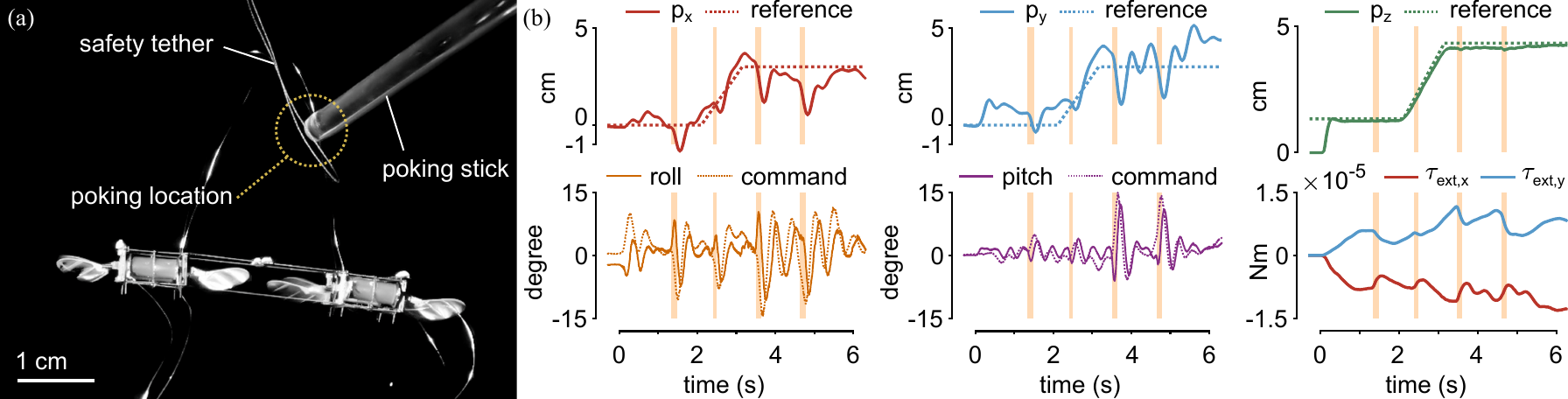}
    \caption{Robustness of the proposed flight control strategy when applying large force/torque disturbances to the \ac{MAV} while tracking a $6.0$ s ramp trajectory on $x$-$y$-$z$, our Task 2 (T2). These results highlight that the robot is not destabilized by the large disturbances, achieving an impressive $0.3$ cm MAE on the $z$ axis. The fast attitude dynamics additionally enable the robot to recover in only $200$ ms. }
    \label{fig:collision}
    \vskip-2ex
\end{figure*}

\subsection{Task 1: Position Ramps}
First, we consider a position ramp of $3$ cm along the three axes of the $W$, with a duration of $1$ s. The total flight time in the experiment is of $3$ s, and we repeat the experiment three times. This is a task of medium difficulty, as the robot needs to simultaneously roll, pitch and accelerate along $z$, but the maneuver covers a small distance ($5.2$ cm). \cref{fig:pos_ramp} (a) shows a time-lapse of the maneuver. \cref{tab:rmse_all} reports the position tracking \ac{RMSE} (computed after take-off, starting at $t_0 = 0.5$), and shows that we can consistently achieve sub-centimeter RMSE on all the axes, with a maximum absolute error (MAE) smaller than $1.6$ cm. This is a $60\%$ reduction over the $4.0$ cm MAE on $x$-$y$ reported in \cite{chen2021collision} for a hover task. Remarkably, the altitude MAE is only $0.2$ cm, with a similar reduction ($60\%$) over \cite{chen2021collision} ($0.5$ cm). \cref{fig:pos_ramp} shows the robot's desired and actual position and attitude across multiple runs (shaded lines), demonstrating repeatability. \cref{fig:pos_ramp} additionally highlights the role of the torque observer, which estimates a position-dependent disturbance, possibly caused by the forces applied by the power cables, or by the safety tether. 

\subsection{Task 2: Rejection of Large External Disturbances}
Next, we increase the complexity of the task by intentionally applying strong disturbances with a stick to the safety tether (\cref{fig:collision} (a)), while tracking the same ramp trajectory in Task 1. This causes accelerations $> 0.25$ g.  \cref{fig:collision} reports position, attitude, and estimated disturbances, where the contact periods have been highlighted in orange. \cref{tab:rmse_all} reports RMSE and MAE. From \cref{fig:collision} (b) we highlight that
\begin{inparaenum}[a)]
\item the position of the robot remains close to the reference ($<2.5$ cm MAE on $x$-$y$) and, surprisingly, the altitude is almost unperturbed ($0.3$ cm MAE).
\item the robot, thanks to its small inertia, recovers quickly from the large disturbances, being capable to permanently reduce (after impact, until the next impact) its acceleration by $50\%$ in less than $200$ ms;
\item despite the impulse-like nature of the applied disturbance, the torque observer detects some of its effects.
\end{inparaenum}

\subsection{Task 3: Circular Trajectory}
\begin{figure}
    \centering
    \includegraphics[width=\columnwidth, trim={0, 8, 0, 0}, clip]{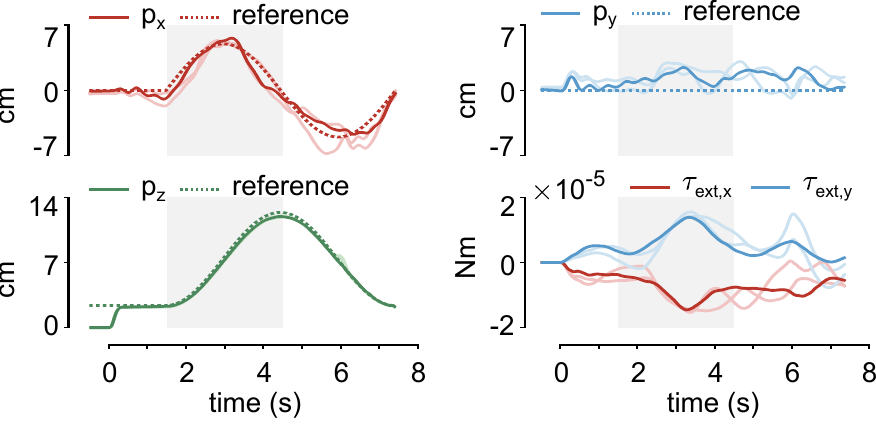} 
    \caption{Performance in Task 3 (T3), where the robot tracks a long ($7.5$ s) circular trajectory, with a wide ($5.0$ cm) radius. The experiment has been repeated three times, and the results are included in the shaded lines. These results highlight the high and repeatable accuracy during agile flight of the \ac{RTMPC}-like neural network policy controlling the robot, as well as the role of the torque observer, which is capable to estimate and compensate for the effects of large rotational uncertainties. A composite image of the trajectory is shown in \cref{fig:teaser}.}
    \label{fig:circular_traj}
    \vskip-2ex
\end{figure}
Last, we track a circular trajectory along the $x$-$z$ axis of $W$ for three times. The trajectory has a duration of $7.5$ s (including takeoff), with a desired velocity of $5.2$ cm/s, and the circle has a radius of $5.0$ cm. A composite image of the experiment is shown in \cref{fig:teaser}. \cref{fig:circular_traj} provides a qualitative evaluation of the performance of our approach, while \cref{tab:rmse_all} reports the \ac{RMSE} of position tracking across the three runs. These results highlight that:
\begin{inparaenum}[a)]
\item in line with Task 1, our approach consistently achieves mm-level accuracy in altitude tracking ($3$ mm average RMSE on $z$), and cm-level accuracy in $x$-$y$ position tracking (RMSE $<1.8$ cm, MAE $<3.1$ cm);
\item the external torque observer plays an important role in detecting and compensating external torque disturbances, which corresponds to approximately $30\%$ of the maximum torque control authority around $\prescript{}{B}{x}$.  
\end{inparaenum}

\section{Conclusions}
\label{sec:conclusions}
This work has presented the first robust MPC-like neural network policy capable of experimentally controlling a sub-gram \ac{MAV} \cite{chen2021collision}. In our experimental evaluations, the proposed policy achieved high control rates ($2$ kHz) on a small offboard computer, while demonstrating small (< $1.8$ cm) RMS tracking errors, and the ability to withstand large external disturbances. 
These results open up novel and exciting opportunities for agile control of sub-gram \acp{MAV}. First, the demonstrated robustness and computational efficiency paves the way for onboard deployment under real-world uncertainties. Second, the newly-developed trajectory tracking capabilities enable data collection at different flight regimes, for model discovery and identification \cite{brunton2016discovering, fasel2022ensemble}. 
Last, as the computational cost of a learned neural-network policy grows favorably with respect to state size, we can use larger, more sophisticated models for control design, further exploiting the nimble characteristics of sub-gram \acp{MAV}.

\balance
\bibliographystyle{IEEEtran}
\bibliography{root}

\end{document}